\pdfoutput=1

\documentclass[11pt]{article}

\usepackage{acl}

\usepackage{times}
\usepackage{latexsym}

\usepackage[T1]{fontenc}

\usepackage[utf8]{inputenc}

\usepackage{microtype}
\usepackage{graphicx}
\title{CMNEROne at SemEval-2022 Task 11: Code-Mixed Named Entity Recognition by leveraging multilingual data}

\author{Suman Dowlagar \\
  LTRC \\
  IIIT-Hyderabad \\
  \texttt{suman.dowlagar}\\
  \texttt{@research.iiit.ac.in} \\\And
  Radhika Mamidi \\
  LTRC \\
  IIIT-Hyderabad \\
  \texttt{radhika.mamidi} \\
  \texttt{@iiit.ac.in} \\}

\begin{document}
\maketitle
\begin{abstract}
Identifying named entities is, in general, a practical and challenging task in the field of Natural Language Processing. Named Entity Recognition on the code-mixed text is further challenging due to the linguistic complexity resulting from the nature of the mixing. This paper addresses the submission of team CMNEROne to the SEMEVAL 2022 shared task 11 MultiCoNER. The Code-mixed NER task aimed to identify named entities on the code-mixed dataset. Our work consists of Named Entity Recognition (NER) on the code-mixed dataset by leveraging the multilingual data. We achieved a weighted average F1 score of 0.7044, i.e., 6\% greater than the baseline.
\end{abstract}

\section{Introduction}
Named entity recognition (NER) is a fundamental task in NLP. It aims to identify and classify entities in a text into predefined types. It is an essential tool for information retrieval, question answering, \cite{banerjee2019information} and text summarization tasks \cite{patil2016survey,li2020survey}. However, except for some resource-rich monolingual languages, NER annotated data for most other languages are still very limited \cite{kruengkrai2020improving}. Moreover, it is usually time-consuming to annotate such data, particularly for low-resource languages such as multilingual and code-mixed \cite{liu2021mulda}. Therefore, transfer learning and leveraging the other datasets for multilingual and code-mixed NER has attracted growing interest recently, especially with the influx of deep learning methods.

This paper presents the system description for named entity recognition on the code-mixed dataset. Code-mixing is defined as using two or more languages in a single sentence or utterance \cite{dowlagar2021offlangone}. The use of code-mixed language is prevalent in most multilingual societies. Due to linguistic complexity arising from mixing two languages, the processing of code-mixed sentences is a challenging task \cite{bali2014borrowing}. So, the models that are trained on monolingual and multilingual datasets typically fail to handle code-mixed inputs \cite{khanuja2020new}. Therefore, to encourage research on code-mixing, the speech and NLP communities are organizing several shared tasks. The shared tasks have concentrated on language identification, POS-tagging, sentiment analysis, hate speech detection, and several datasets exist for these as well. Similarly, SEMEVAL 2022's Task 11 sub-task 13 was devoted to identifying named entities in code-mixed languages \cite{multiconer-report}. This task aims to classify the given tokens in the code-mixed sentences as persons, corporation, location, and others. An example is shown in Table \ref{tab:example1}.

\begin{table}[]
\begin{tabular}{|l|llll|}
\hline
\textbf{Sentence} & \textit{hameM} & this & magic & moment \\
\hline
\textbf{Languge} & Hi & En & En & En \\
\hline
\textbf{NER tags} & O & B-CW & I-CW & I-CW \\
\hline
\end{tabular}
\caption{An example of the code-mixed NER annotated sentence. The Hindi words are converted from utf to wx format and are italicized}
\label{tab:example1}
\end{table}

\begin{table*}[]
\begin{tabular}{llllllllll}
\hline
\textit{sIriyala} & \textit{naMbara} & \textit{xvArA} & kleding & in & de & oudheid & \textit{kI} & \textit{pahacAna} & \textit{kareM} \\
O & O & O & B-PROD & I-PROD & I-PROD & I-PROD & O & O & O \\
\hline
what & city & is & dig & me & out & in? &  &  &  \\
O & 0 & O & B-CW & I-CW & I-CW & O &  &  &  \\
\hline
\textit{AmAra} & das & testament & \textit{mUlya} & \textit{kawa?} &  &  &  &  &  \\
0 & B-CW & I-CW & O & O &  &  &  &  & \\
\hline
\end{tabular}
\caption{An example of the code-mixed NER annotated sentence. The multilingual words are converted from utf to wx format and are italicized}
\label{tab:example2}
\end{table*}

The lack of annotated data is a crucial issue for code-mixed datasets. Lack of data poses a problem of data overfitting and poor entity recognition. The language models trained on such low resource datasets cannot generalize the training data, thus performing low on the test datasets. Several previous studies have used monolingual data as training signals for transfer learning, and these data can also be used in the form of pre-training. Thus, we used a similar approach of including the multilingual data along with the code-mixed dataset.

We used the multilingual pre-trained BERT model as our model for code-mixed NER. The model uses code mixed training data along with the multilingual training and mulilingual validation data.

We have analyzed that Bi-LSTM with CRF models has shown an improved accuracy on the token classification tasks such as POS tagging, language identification, and NER. The ensemble of BERT or XLM-RoBERTa with Bi-LSTM and CRF would have shown a further improvement in the code-mixed NER. Also, using language identification as a downstream task with the current method might have improved the NER's accuracy.

The paper is organized as follows. Section \ref{rw} provides related work on Named Entity Recognition on CM social media text. Section \ref{data} provides information on the task and examples of datasets. Section \ref{work} describes the proposed work. Section \ref{experimental_setup} presents the experimental setup and Section \ref{results} project the performance of the model. Section \ref{conclude} concludes our work.

\section{Related Work}
\label{rw}
Code-mixed NER has attracted a lot of attention in the NLP community this decade. This section lists the latest works on code-mixed named entity recognition.  

\citet{priyadharshini2020named,winata2019learning} generated multilingual meta representations from pre-trained monolingual word embeddings. The model learned to construct the best word representation by mixing multiple sources without explicit language identification.

\citet{aguilar2019named} presented a shared task on named entity recognition in the CALCS workshop. The language pairs used were English-Spanish (ENG-SPA) and Modern Standard Arabic Egyptian (MSA-EGY). They used Twitter data and nine entity types to establish a new dataset for code-switched NER benchmarks. The participating teams used LSTM, CNN, CRF, and word representations to recognize named entities.

\citet{singh2018named} We presented a new Code-Mixed Hinglish corpus for NER. Different machine learning classification algorithms with word, character, and lexical features are used to establish baselines. The algorithms used were Decision tree, Long Short-Term Memory (LSTM), and Conditional Random Field (CRF).

\cite{meng2021gemnet,fetahu2021gazetteer} presented a novel CM NER model. They proposed a gated architecture that enhances existing multilingual Transformers by dynamically infusing multilingual knowledge bases, a.k.a gazetteers. The evaluation of code-mixed queries shows that this approach efficiently utilizes gazetteers to recognize entities in code-mixed queries with an F1=68\%, an absolute improvement of +31\% over a non-gazetteer baseline. 

\cite{meng2021gemnet} mentioned that including Gazetteer features could cause models to overuse or underuse them, leading to poor generalization. They proposed a new approach for gazetteer knowledge integration by including Context in Gazetteer Representation using encoder and Mixture-of-Experts gating network models. These models overcome the feature overuse issue by learning to conditionally combine the context and gazetteer features instead of assigning them fixed weights. 

\section{Task Setup}
\label{data}
The shared task detects semantically ambiguous and complex entities in short and low-context settings. Complex NEs, like the titles of creative works (movie/book/song/software names), are not simple nouns. Usually, they take imperative clauses, or they often resemble typical syntactic constituents. Such NEs are harder to recognize \cite{ashwini2014targetable}. Syntactic parsing of such complex noun phrases is hard, and most NER systems fail to identify them. Inside–outside–beginning (IOB) format \cite{ramshaw1999text} is used for annotating entities. A few examples of complex NEs and ambiguous NEs from the code-mixed dataset are given in Table \ref{tab:example2}.

So the MultiCoNER shared task encourages the models to handle such complex NEs. A huge dataset \cite{multiconer-data} is released for this task \cite{multiconer-report}. The languages focused on in this shared task are: English, Spanish, Dutch, Russian, Turkish, Korean, Farsi, German, Chinese, Hindi, and Bangla. The shared task also offered an additional track with code-mixed and multilingual datasets. In this paper, we will be concentrating on the code-mixed dataset.

\section{System Overview}
\label{work}
We finetuned the pre-trained multilingual BERT model by using the multilingual training and validation datasets for code-mixed named entity recognition. We found that the training data is insufficient for the deep learning language model to identify the named entities in the validation data correctly. The data scarcity of low-resource languages has been a significant challenge for building NLP systems since they require a large amount of data to learn a robust model. We observed that the multilingual NER training data is similar to the code-mixed dataset. Also, it is relatively large when compared to the code-mixed dataset. In our approach, the multilingual training and validation data is combined with the code-mixed training dataset. Using the combined dataset, we finetune the deep neural network model. Our method thus attempts to learn language-agnostic features by using the combined multilingual and code-mixed dataset. This finetuned model can be used to infer named entity information at a token level on a code-mixed low resource language.

We used the pre-trained mBERT \cite{devlin2018bert} model for code-mixed NER. mBERT is a transformer-based self multi-headed attention model that is pre-trained on a massive collection of multilingual data and can be finetuned for our NER task. As the model is pre-trained on a large corpus, the semantic and syntactic information is well modeled and can be directly finetuned for a specific task. BERT is a bi-directional transformer model \cite{vaswani2017attention}. It analyzes the meaning of a term depending on its context given on both sides. The transformer part in the BERT works like an attention mechanism capable of learning the contextual relationships between the terms in a sentence.

\section{Experimental Setup}
\label{experimental_setup}
The section presents the baselines, hyper-parameter settings, and analysis of observed results. 

\subsection{Baselines}

The baselines used for the proposed work is:

\paragraph{Conditional random field (CRF) \cite{lafferty2001conditional}} CRF is a statistical model and is a well-known approach for handling NER problems. The CRF model considers the neighboring samples by modeling the prediction as a graphical model. It assumes that the tag for the present word (denoted as $y_{i}$) is dependent on the tag of its previous/next word (denoted as $y_{i-1}$ or $y_{i+1}$). 

\paragraph{MultiCoNER baseline \cite{multiconer-report}} The XLM-RoBERTa base with CRF model is used as a baseline for NER.

\paragraph{Pre-trained multilingual BERT (mBERT) \cite{devlin2018bert}} A pre-trained multilingual BERT model with token classification without leveraging the multilingual data is used as a baseline.

\subsection{Hyperparameters and libraries}

For developing our model, the neural network library used is PyTorch, and the pre-trained multilingual BERT model (\textit{bert-base-multilingual-cased}) and XLM-ROBERTa base model (\textit{xlm-roberta-base}) is obtained from the hugging face-transformers library and is finetuned for the code-mixed NER task. The model is implemented in Kaggle Notebook with GPU processing. 

The batch size of the datasets is kept as 64. The maximum length of the sentence from the training data is considered during the input data encoding/padding. Due to subword tokenization, we used the first token for predicting the tag. The optimizer used is weighted Adam with the learning rate of 2e-5 and epsilon value equal to 1e-5. The dropout is set to 0.1. The loss function used is a cross-entropy loss that is inbuilt into the transformer's BERT model. The number of epochs used for training the model is 30. The training is stopped when there is no change in validation accuracy for more than four epochs.

\begin{table*}[]
\centering
\begin{tabular}{|l|ll|ll|}
\hline
\textbf{Model} & \multicolumn{4}{c|}{\textbf{F1-score}} \\ \cline{2-5} 
 & \multicolumn{2}{l|}{\textbf{w/o multilingual data}} & \multicolumn{2}{l|}{\textbf{with multilingual data}} \\ \cline{2-5} 
 & \textbf{valid data} & \textbf{test data} & \textbf{valid data} & \textbf{test data} \\ \hline 
CRF & 0.565 & 0.561 & 0.560 & 0.556 \\
mBERT & 0.627 & 0.612 &\textbf{0.716} & \textbf{0.707} \\
MultiCoNER baseline & 0.651 & 0.645 & 0.725 & 0.719 \\
\hline
\end{tabular}
\caption{The performance of the models on the code-mixed dataset with and without including multilingual data. Our submission for the given task is highlighted.}
\label{tab:result}
\end{table*}

\begin{figure}[h]
\includegraphics[width = \columnwidth]{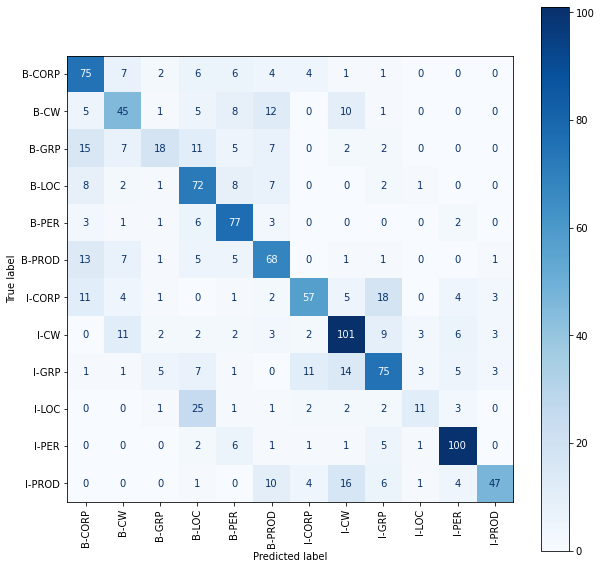}
\caption{Confusion matrix of CM-NER baseline}
\label{cm1}
\end{figure}

\begin{figure}[h]
\includegraphics[width = \columnwidth]{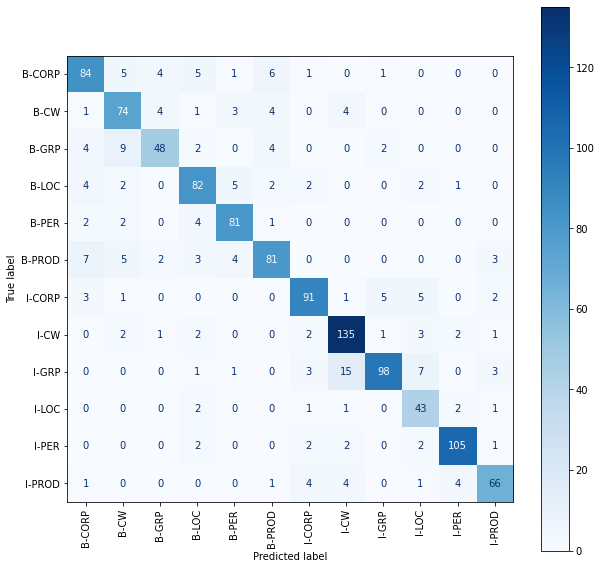}
\caption{Confusion matrix of CM-NER by leveraging multilingual data}
\label{cm2}
\end{figure}

The CRF is obtained from pytorch-crfsuite library\footnote{https://github.com/scrapinghub/python-crfsuite}. The previous word and its tag, the next word, and its tag is used as the features to predict the tag of the current word.

\section{Results and Analysis}
\label{results}
Table \ref{tab:result} presents the f1-score of the models on the Dravidian code-mixed dataset. From the above results, it is clear that our system, i.e., leveraging the multilingual NER data in a low-resource code-mixed setting, improves the NER task compared to the baseline models. The CRF model didn't perform well on the given NER task, as this statistical model does not capture the semantics of the tokens. Even the CRF with multilingual data performed poorly on this task compared to the baseline NN models. It shows the importance of capturing semantical, syntactic, and contextual information while building the NER model on these complex datasets. 

Our submission, the pre-trained mBERT by leveraging the multilingual dataset, performed better than the MultiCoNER baseline by 6\%. Even the MultiCoNER baseline with the multilingual dataset performed better than our submission. 

The confusion matrices with and without multilingual data of our submission on the code-mixed NER validation dataset are shown in the Figures \ref{cm1} and \ref{cm2}. By using confusion matrices, we observed that the multilingual data given in Figure \ref{cm2} helped better identify the  CW, PROD, CROP, and LOC entities when compared to the baseline model.

\section{Conclusion and future work}
\label{conclude}
In this paper, we addressed the shared task on named entity recognition for the code-mixed dataset. As the code-mixed data is a low resource language and there are no pre-trained models, we leveraged the multilingual dataset for training the NER model. The model used for testing our method is the pre-trained multilingual BERT. We finetuned the pre-trained mBERT for the code-mixed NER task by using the code-mixed training data and multilingual training and validation datasets. 

The use of meta embeddings for dealing with code-mixed datasets has recently attracted a lot of attention. It might be possible that meta embedding-based NER will work better on this code-mixed dataset. Unlike the social media data where code-mixed sentences/words are written in Roman script, the native script is used for each word, so the language identification will work better on this dataset. Using Language identification or POS tagging as a downstream task for NER on this dataset will help in improving the code-mixed NER.

\bibliography{anthology,custom}
\bibliographystyle{acl_natbib}

\end{document}